\documentclass[runningheads]{llncs}

\usepackage{eccv}

\usepackage{eccvabbrv}

\usepackage{graphicx}
\usepackage{booktabs}
\usepackage{times}
\usepackage{amsmath, amssymb}
\usepackage{indentfirst}
\usepackage{subcaption}
\usepackage{multirow}
\usepackage{bm}
\usepackage[accsupp]{axessibility}  % Improves PDF readability for those with disabilities.

\usepackage{hyperref}

\usepackage{orcidlink}

\begin{document}

\title{Perceiving Better Moments: Cover Frame Reselection and Enhancement for Live Photos with the Live2K Dataset}

\titlerunning{Perceiving Better Moments}

\author{
Junyu Lou\inst{1} \and
Kai Chen\inst{1} \and
Weiyi You\inst{1} \and
Hui Zeng\inst{2} \and
Lei Zhang\inst{2,3} \and
Shuhang Gu\inst{1}\thanks{Corresponding author}
}

\authorrunning{J. Lou et al.}

\institute{
University of Electronic Science and Technology of China
\and
OPPO Research Institute
\and
The Hong Kong Polytechnic University\\[0.5em]
\email{\{junyulou.jy, shuhanggu\}@gmail.com}\\
\url{https://github.com/CVL-UESTC/Live2K}
}

\maketitle

\begin{abstract}
Modern smartphones capture Live Photos—short video bursts surrounding a still image—offering a dynamic and engaging photographic experience. However, the cover photo and video components are generated by two distinct imaging pipelines: the photo stream undergoes full computational photography processing, while the video stream is constrained by real-time efficiency and heavy compression. This intrinsic separation produces a substantial quality gap in resolution, color fidelity, and dynamic range between the cover photo and video frames.
When users reselect an alternative frame from the video to replace an imperfect cover, the chosen frame often suffers from severe degradation, making direct replacement visually unsatisfactory. Restoring such frames requires simultaneous enhancement of spatial detail and color appearance, a task considerably more challenging than ordinary super-resolution or color enhancement.
To address this, we define the Live Photo Cover Frame Reselection and Enhancement (LPRE) task, which leverages the intrinsic cues available within each Live Photo: the high-quality cover image as a structural and color reference, the user-reselected low-quality frame as the reconstruction target and several adjacent video frames providing temporal cues. Building upon this formulation, we construct Live2K, a real-world dataset of 2,042 Live Photos, and develop a unified one-stage baseline that integrates multi-frame fusion, guided color enhancement and super-resolution—establishing the first benchmark for Live Photo enhancement research.
  \keywords{Live Photo \and Image Enhancement \and Image Super-Resolution}
\end{abstract}

\section{Introduction}
\label{sec:intro}

\begin{figure}[t]
  \flushright
  \includegraphics[width=1\linewidth]{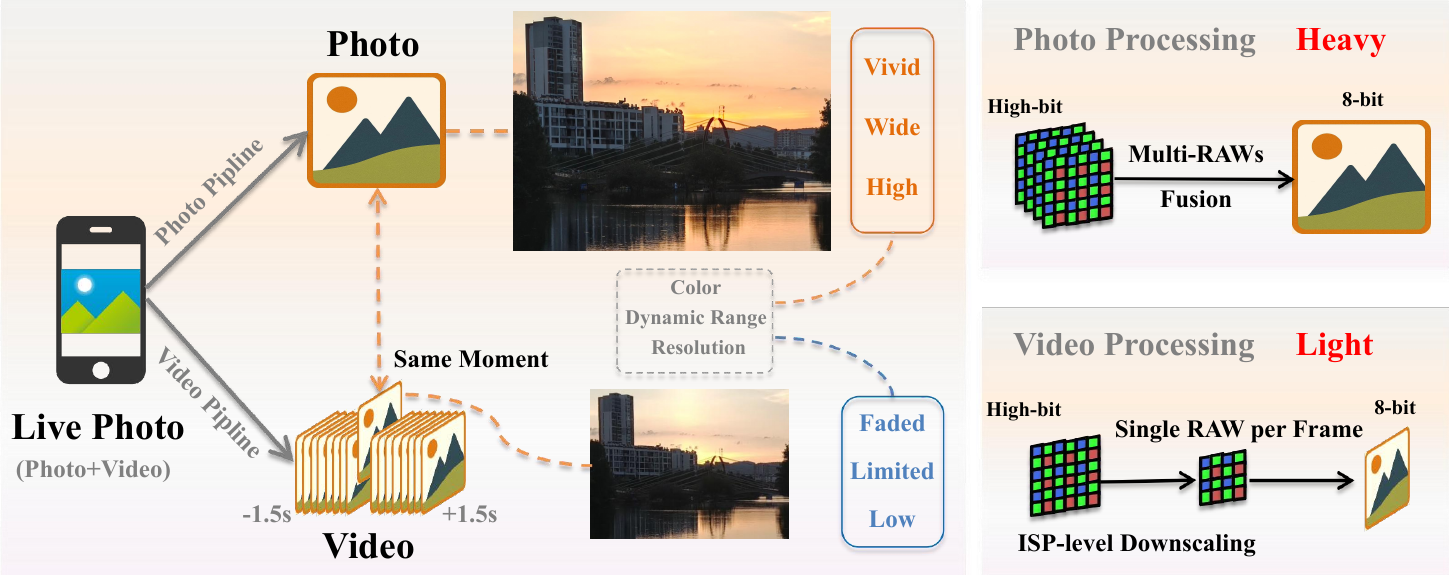}
  \caption{Comparison between the cover frame (top) and the video frame (bottom) captured at the same moment in Live Photos. The \textcolor{red}{photo pipeline} processes \textcolor{red}{multiple} high-bit RAW data with \textcolor{red}{rich} system resources, resulting in a high-quality cover frame;
meanwhile, the \textcolor{blue}{video pipeline} processes each video frame with a \textcolor{blue}{single} RAW image under a \textcolor{blue}{tight} computational budget, resulting in low-quality video frames.}
  \label{fig:figure1}
\end{figure}

Modern smartphones are typically equipped with the feature of Live Photos. When capturing a still image, the device simultaneously records a short video clip encompassing moments before and after the shutter is pressed, thereby creating a brief “living moment.” During playback, this produces a more immersive and narrative experience than a conventional photo, making Live Photos widely popular among users. 

\begin{figure*}[t]
  \centering
  \includegraphics[width=\textwidth]{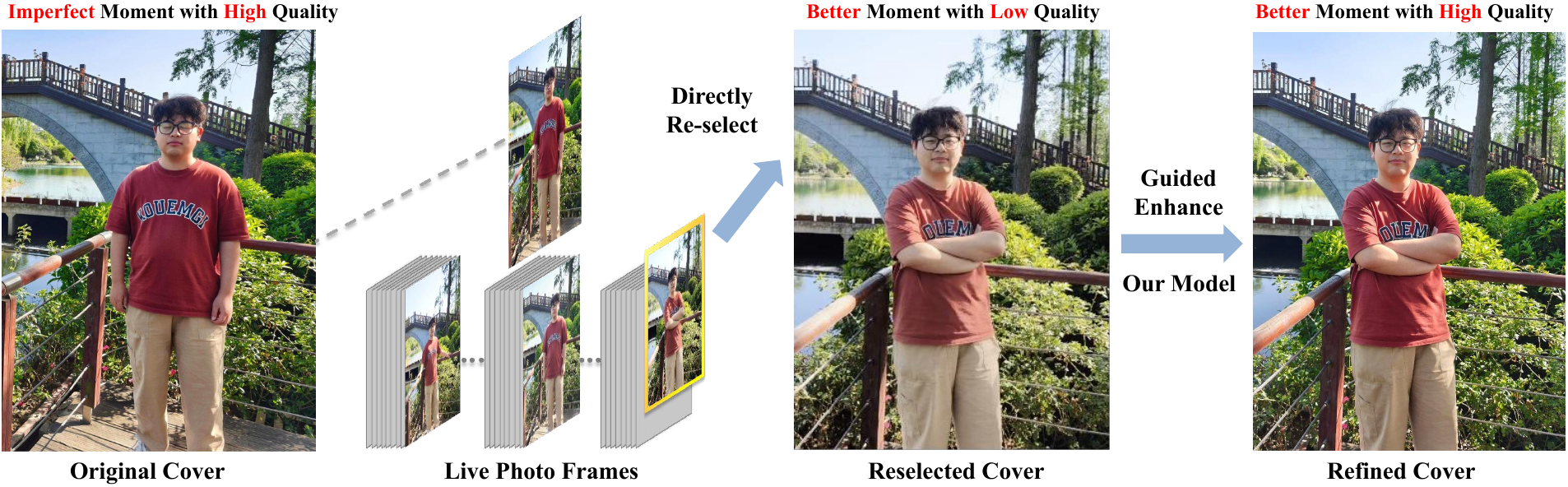}
  \caption{When users attempt to replace an unsatisfactory cover, the newly selected cover frame is often of inferior quality. Therefore, a model is needed to perform comprehensive enhancement to restore and improve its visual quality.}
  \label{fig:figure2}
\end{figure*}

The photo and video components of a Live Photo are produced by two distinct imaging pipelines. The photo stream (used for the cover image) is generated through multi-frame fusion of several high–bit-depth RAW images captured at different exposures (long, short, and normal) to achieve a higher dynamic range and superior image quality. The video stream, in contrast, is based on a single RAW frame per time step, as is typical for real-time recording. In the Live Photo mode, the still photo and video must be captured simultaneously, which imposes strict latency and storage constraints on the video pipeline. As a result, the budget allocated to each frame is very limited. To meet these requirements, the video frames are typically downsampled at an early stage of processing and then processed with a lightweight, efficiency-oriented pipeline. As depicted in Fig.~\ref{fig:figure1}, this fundamental separation leads to an intrinsic quality gap between the cover image and video frames—manifested primarily in resolution, color fidelity, and dynamic range. These discrepancies stem from the imaging system itself and cannot be eliminated due to the system’s latency demands. This issue often undermines user experience, as the automatically captured cover photo may correspond to an imperfect moment—such as a blink or an unideal expression—prompting users to reselect another frame from the video. Yet the reselected frame typically exhibits lower resolution, faded colors, and limited dynamic range, resulting in a visible degradation compared to the original cover, as shown in Fig.~\ref{fig:figure2}. Making this re-selection viable thus requires a model capable of restoring structural detail while correcting appearance inconsistencies in tone. In other words, the task demands simultaneous enhancement of both spatial resolution and color quality, which makes it more challenging than ordinary color enhancement or super-resolution.

\label{sec:dataset}

Fortunately, every Live Photo inherently contains a high-quality still image—the original cover—that shares strong content similarity with the reselected frame. This provides a powerful cue for guided enhancement, where the cover image serves as a reference for recovering fine details and accurate color rendition. Moreover, the neighboring frames in the video sequence offer complementary temporal cues that further aid the reconstruction. Together, these observations motivate a new and practical problem: the Live Photo Cover Frame Reselection and Enhancement (LPRE) task, which aims to transform a user-selected low-quality frame into a high-quality photo consistent in visual appearance with the original cover.

To enable research on this problem, we construct Live2K, a large-scale collection of 2,042 real-world Live Photos captured using two smartphone systems. Each Live Photo contains a high-quality cover image and its accompanying short video sequence. Building such a collection is non-trivial, as Live Photos are device-specific and their internal imaging pipelines are proprietary.
Based on these raw Live Photos, we further construct paired data for the reselection and enhancement task. Each final pair consists of one guidance image, one ground-truth image, one corresponding low-quality frame from the video sequence, and several neighboring frames as temporal context. Importantly, the guidance image is drawn from another Live Photo captured under a visually similar scene, providing color and texture priors from a distinct yet related sample. 
To our knowledge, Live2K is the first dataset specifically designed for the Live Photo reselection and enhancement task. Based on Live2K, we design a unified one-stage baseline that jointly performs multi-frame fusion, guided enhancement, and super-resolution, supervised by intermediate color loss and final reconstruction loss. It is worth noting that instead of pursuing architectural novelties, our primary objective is to establish a solid and unified baseline for the newly introduced LPRE task.

Our main contributions are summarized as follows:
\begin{itemize}
    \item[$\bullet$] We introduce the Live Photo Cover Frame Reselection and Enhancement (LPRE) task, aimed at bridging the intrinsic quality gap between cover and video frames.
    \item[$\bullet$] We construct Live2K, the first large-scale real-world dataset tailored for this new task.
    \item[$\bullet$] We present a unified baseline framework that integrates temporal fusion and guided enhancement, offering the first benchmark for Live Photo enhancement research.
\end{itemize}

\section{Related Work}
\label{sec:related}

\subsection{Image Enhancement}

Traditional image enhancement approaches rely on handcrafted priors such as histogram equalization~\cite{pizer1987adaptive,abdullah2007dynamic} or Retinex-based decomposition~\cite{rahman1996multi,land1971lightness}, which are often inadequate under complex illumination and device-dependent degradations. Deep learning has largely replaced these rule-based designs, learning content-adaptive color and tone mappings from data. Representative works include bilateral grid processing~\cite{HDRNet,lou2025learning} and LUT-based models~\cite{3DLUT,AdaInt,SA3DLUT} as well as curve regression methods~\cite{ZeroDCE}, achieving real-time performance while preserving perceptual quality. 

Some reference-based studies~\cite{PhotoWCT,luan2017deep,gatys2016image,ke2023neural} explore photorealistic color transfer, where an example provides guidance for global tone alignment. However, these works primarily pursue stylistic consistency rather than faithful enhancement. In contrast, our approach leverages the high-quality cover frame in Live Photos as a reliable prior that supplies contextual and structural cues, supplementing the degraded frame to achieve more accurate and natural enhancement under cross-quality conditions.

\subsection{Image Super-Resolution}
Single-image super-resolution (SISR) reconstructs a high-resolution image from a single low-resolution input.
Early CNN-based methods established the basic paradigm~\cite{dong2015image}, followed by deeper residual~\cite{lim2017enhanced,kim2016accurate, zhang2018residual}, attention-based~\cite{zhang2018image}, and Transformer architectures~\cite{liang2021swinir,zhang2024transcending,chen2023activating,long2025progressive}.
Although effective, SISR is inherently limited by the lack of complementary information within a single frame.

To overcome this, burst and multi-frame super-resolution approaches exploit short-term temporal redundancy from multiple observations.
Burst SR~\cite{bhat2021deep,dudhane2022burst} focuses on handheld or smartphone bursts, aligning frames with sub-pixel precision to aggregate fine details,
while video super-resolution approaches~\cite{chan2021basicvsr,wang2019edvr,kappeler2016video,zhou2024video} extends this idea to longer sequences with motion compensation and temporal alignment.
These methods demonstrate significant improvements in reconstruction quality but remain challenged by motion variations and illumination inconsistencies.

Reference-based super-resolution approaches methods introduce an additional high-quality image to provide texture or structural priors~\cite{jiang2021robust,lu2021masa,zhang2019image,yang2020learning,cao2022reference,kim2023efficient,lee2022reference}.
By establishing feature-space correspondence, these methods can transfer fine details and recover high-frequency content when reliable guidance is available.

Our pipeline combines multi-frame inputs with cross-quality guidance in Live Photos.
The short burst sequence provides temporal redundancy, while the fully processed cover frame offers reliable color and structural priors.
Our method unifies these perspectives by performing correspondence-aware fusion that jointly exploits multi-frame complementarity and high-quality guidance under cross-device conditions.

\section{Live2K Dataset}
\label{sec:dataset}
\begin{figure}[t]
  \flushright
  \includegraphics[width=1\linewidth]{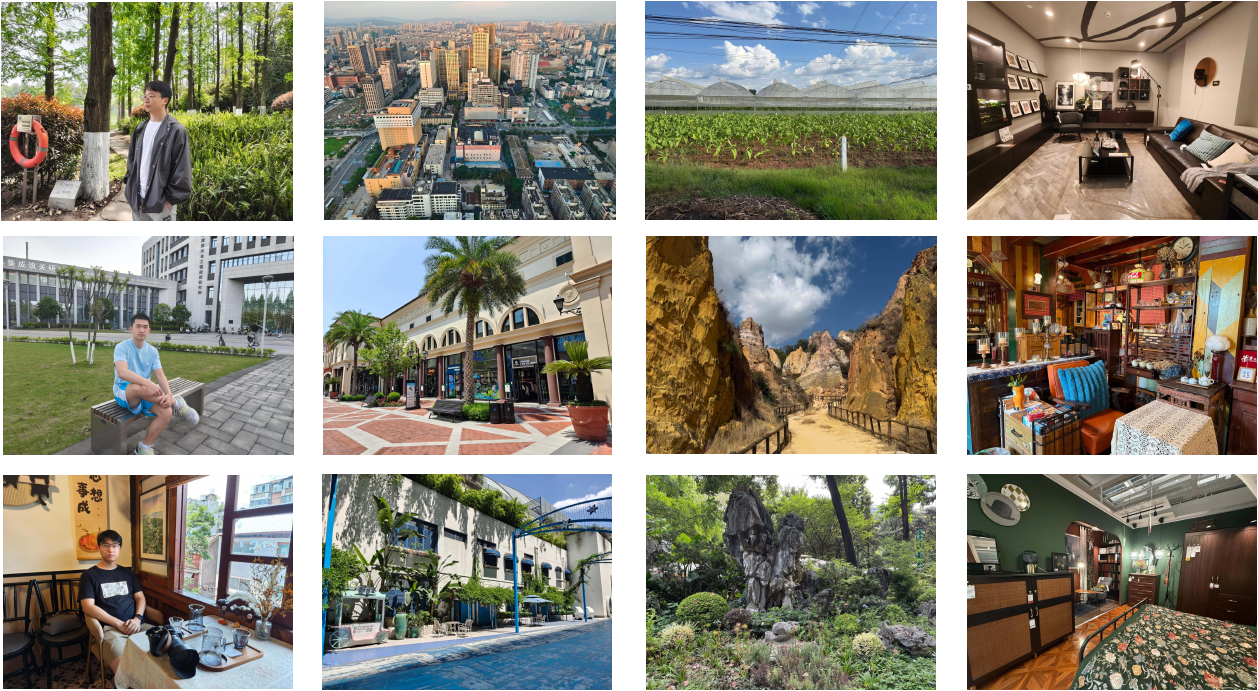}
  \caption{Example ground truth images in the Live2K dataset. Each column corresponds to a category: portrait, architecture, landscape, and indoor scene, from left to right.}
  \label{fig:figure3}
\end{figure}
We construct a real-world Live Photo dataset, named \textbf{Live2K}, to support the training and evaluation of the reselected frame enhancement task.

\subsection{Data Capture with Smartphones}
We use an iPhone 16 Pro and an OPPO Find X8 Pro to capture 2,042 real-world Live Photos, each consisting of a cover image and its corresponding video sequence. The iPhone contributed 1,021 samples, with cover images at a resolution of 4032×3024 and video frames at 1920×1440, captured using its fusion camera system with a 24 mm f/1.78 lens. The OPPO contributed 1,021 samples, with cover images at 4096×3072 and video frames at 1728×1296, captured using a 23 mm f/1.6 lens. Since the two devices have distinct ISP pipelines and super-resolution scales, we conducted separate training and testing on the datasets from each device. 
The dataset covers a wide range of scenes, including portraits, architecture, natural landscapes, and indoor scenes. A small sample of images is shown in Fig.~\ref{fig:figure3}, which showcases the diversity of our dataset.  

 To provide data suitable for cross-instance guidance, the dataset was collected in a group-wise manner, where each group contains multiple Live Photos with similar content, allowing the covers to serve as mutual guidance images. Live Photos within the same group were taken from the same scene, but with variations in camera angle, distance, scene composition, and subject movement to obtain visually similar yet non-identical images. This setup allows the guidance images and video frames to closely mimic the real-world relationship between the cover and video frames in actual Live Photo captures.

\subsection{Data Preprocessing}
To construct data pairs, we need to find the video frame that best corresponds to the cover image at the pixel level. For each Live Photo, we first downsample the cover image to match the resolution of the video frames and convert both the cover and video frames into the grayscale domain, which reduces color differences and allows the matching to focus on structural cues. We then use the SIFT~\cite{lowe1999object} algorithm to identify the video frame most similar to the cover image, which serves as its corresponding low-quality counterpart. Since Live Photos capture video footage extending 1.5 seconds before and after the shutter press, the most similar frames are consistently identified near the median position of the sequence. Given that slight viewpoint differences may exist between the cover and the matched frame, we further warp the cover image to align with the frame’s perspective using a homography estimated from SIFT feature correspondences, obtaining the ground-truth image. We also select the eight neighboring frames (four before and four after) around the matched frame, forming a 9-frame input sequence. The dataset is organized into 751 groups, in the training set, each group contains three images that can guide each other, resulting in six paired samples per group. In the test set, each group has two images, forming one paired sample. Overall, the final paired dataset includes 1{,}620 training images (yielding 3{,}240 training pairs) and 422 test images (yielding 211 test pairs).

We adopt patch-based training to reduce GPU memory usage and improve computational efficiency. For each data pair in the training set, we first perform feature matching between the ground truth (GT) and the reference (Ref) images using the SIFT~\cite{lowe1999object} algorithm. Then, a sliding-window search is applied to locate, for each GT patch, the most similar patch in the Ref image. Based on the desired super-resolution scale, the corresponding region in the input frames is extracted to form a new paired sample. To ensure reliable correspondence, we apply a similarity threshold to the matched feature points, and pairs below this threshold are discarded. The thresholds are set to 160 for iPhone and 120 for OPPO. 
As a result, the iPhone subset contains 60{,}291 training pairs, where both GT and Ref images are of size 504×504 and the input images are 240×240, while the OPPO subset contains 57{,}755 training pairs with GT/Ref sizes of 512×512 and input size of 216×216.

\subsection{Misalignment Analysis}

Real-world datasets~\cite{zhang2019zoom,bhat2021deep,cai2019toward,yang2021real} often suffer from spatial misalignment. Prior studies typically acquire inputs and ground truths (GT) using different sensors at different times, resulting in significant spatial displacement. However, leveraging the unique configuration of Live Photos, we can obtain GT and nearly simultaneous video frames captured by the same sensor. Consequently, the issue of data misalignment is substantially alleviated. We observe from the experimental results that high reconstruction quality can be achieved without any auxiliary alignment techniques.

\section{Guided Enhancement Network}
\label{sec:method}
\begin{figure*}[t]
  \centering
  \includegraphics[width=\textwidth]{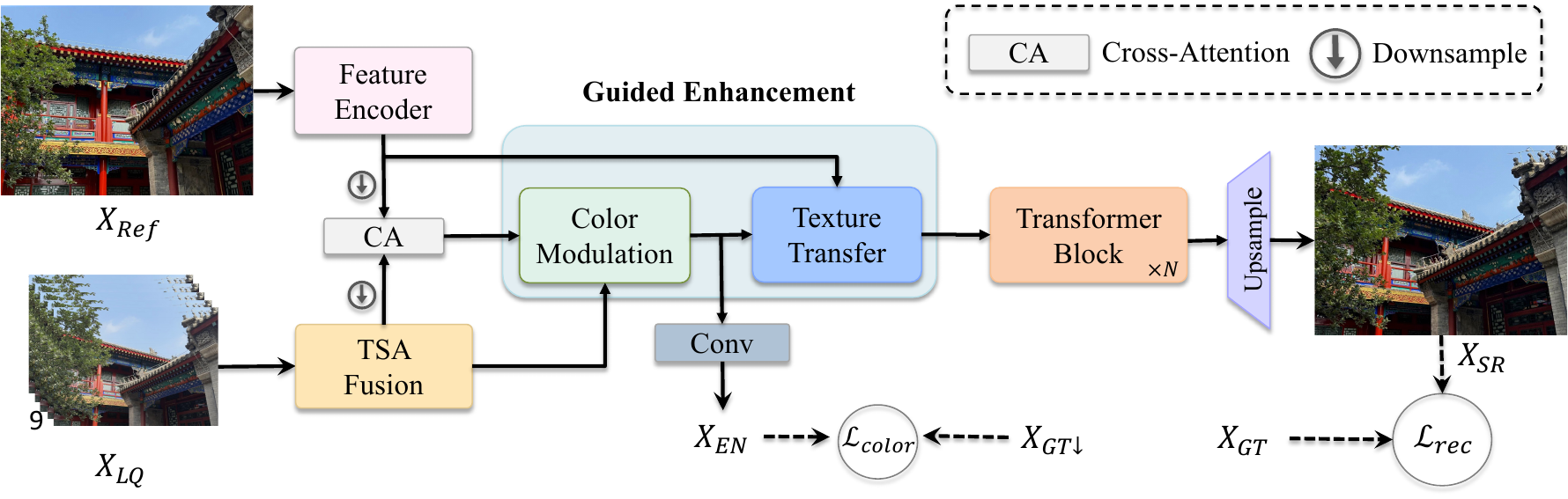}
  \caption{The overall architecture of proposed Live Photo Enhancement Network (LPENet). The Color Modulation module consists of several Guided Color Enhancement Blocks (Fig.~\ref{fig:figure5}).}
  \label{fig:figure4}
\end{figure*}

Given a short burst of consecutive video frames $\{\bm{X}_t\}_{t=1}^{9}\in\mathbb{R}^{3\times h\times w}$ from a Live Photo and a high-quality cover image $\bm{X}_{ref}\in\mathbb{R}^{3\times H\times W}$ as the reference, our goal is to enhance any selected video frame to a visually comparable quality to $\bm{X}_{ref}$. To this end, we propose \textbf{LPENet} (Live Photo Enhancement Network), as illustrated in Fig.~\ref{fig:figure4}.
By integrating temporal fusion and guided enhancement into a streamlined pipeline, we aim to provide a stable foundation to facilitate future research in this domain.

\subsection{Feature Extraction and Multi-Frame Fusion}
To reduce GPU memory consumption, each input frame is first rescaled to half the resolution of the guidance image, followed by a 2× downsampling using the PixelUnshuffle operation. While all subsequent processing is performed at this reduced resolution, performing explicit optical flow-based alignment across 9 frames remains computationally prohibitive. Therefore, instead of relying on optical flow, we employ the Temporal-Spatial Attention (TSA) module adapted from EDVR~\cite{EDVR} to effectively exploit temporal redundancy and handle inter-frame variations. This module is tasked with implicitly aligning and aggregating features from the 9-frame input sequence. By computing attention maps across both temporal and spatial dimensions, the TSA module dynamically fuses vital contexts from neighboring frames, yielding a comprehensive fused feature representation $\bm{F}_{lq}\in\mathbb{R}^{C\times \frac{H}{4}\times \frac{W}{4}}$. Parallel to this, the Feature Encoder extracts texture information from the reference image. It comprises several convolutional layers followed by two PixelUnshuffle operations. This process maps the reference image into a deep embedding space, generating the reference feature map $\bm{F}_{Ref}\in\mathbb{R}^{C\times \frac{H}{4}\times \frac{W}{4}}$ that matches the resolution of the fused video features.

\subsection{Guided Enhancement}
\label{subsec:guided_enhancement}
To achieve color-consistent enhancement and spatial detail reconstruction guided by a high-quality reference, we design a two-stage Guided Enhancement module operating on the feature pair $(\bm{F}_{LQ}, \bm{F}_{Ref})$. Adopting a coarse-to-fine strategy, this module progressively refines the low-quality input by first aligning the overall color through color modulation, and subsequently injecting local details via texture transfer.

\subsubsection{Color Modulation.}
\label{subsubsec:color_modulation}
In the first stage, we introduce a Color Modulation module to produce a color-corrected representation $\bm{F}_{EN}$. The module consists of multiple color enhancement blocks adapted from the NAFBlock~\cite{chen2022simple} architecture, integrated with additional cross-attentive guidance and Feature-wise Linear Modulation (FiLM)~\cite{perez2018film}, as illustrated in Fig.~\ref{fig:figure6}. Since color enhancement depends more on global context than on local texture details, we first perform a cross-attention operation on the downsampled features $(\bm{F}_{LQ\downarrow}, \bm{F}_{Ref\downarrow})$ outside the enhancement blocks to obtain a global guidance feature $\bm{F}_G$. The resulting $\bm{F}_G$ is then upsampled back to the resolution of the input to align with the enhancement backbone. This design not only reduces computational cost but also enables the network to capture global color and tone correspondences between the low-quality input and the reference. The cross-attention is formulated as:
\begin{equation}
\text{Attn}(\bm{Q}, \bm{K}, \bm{V}) = \text{Softmax}\left(\frac{\bm{QK}^{\top}}{\sqrt{d}}\right)\bm{V},
\end{equation}
where $d$ is the head dimension, $\bm{Q}$ is derived from $\bm{F}_{LQ\downarrow}$, and $\bm{K}$ and $\bm{V}$ are obtained from $\bm{F}_{Ref\downarrow}$.

The attended feature $\bm{F}_G$ serves as a shared global guidance signal for all enhancement blocks, ensuring consistent color mapping across the network.
Within each block, $\bm{F}_G$ is further processed by a shallow convolutional layer to produce block-specific modulation maps:
\begin{equation}
[\beta, \gamma] = \text{Conv}(\bm{F}_G), \quad
\beta = \tanh(\beta), \quad
\gamma = \sigma(\gamma),
\end{equation}
where $\tanh(\cdot)$ and $\sigma(\cdot)$ denote the hyperbolic tangent and sigmoid functions, respectively.
The resulting modulation parameters $\beta$ and $\gamma$ are applied to the block’s enhancement backbone in a FiLM-like manner:
\begin{equation}
\begin{aligned}
y = x + f_1(x) \odot \beta, \quad z = y + f_2(y) \odot \gamma,
\end{aligned}
\end{equation}
where $f_1(\cdot)$ and $f_2(\cdot)$ are two residual branches.

This design integrates global color guidance via low-resolution cross-attention and local refinement via FiLM modulation, achieving visually consistent enhancement guided by $\bm{F}_{Ref}$.
To ensure that this module focuses primarily on color transformation and reduces the influence of color discrepancies during subsequent feature alignment, we append a convolutional layer at the end of this module to map $\bm{F}_{EN}$ back to the image domain.
The output image $\bm{X}_{EN}$ is then compared with the downsampled ground truth $\bm{X}_{GT\downarrow}\in\mathbb{R}^{3\times \frac{H}{4}\times \frac{W}{4}}$ to compute a color-related loss.

\begin{figure*}[t]
  \centering
  \includegraphics[width=0.7\textwidth]{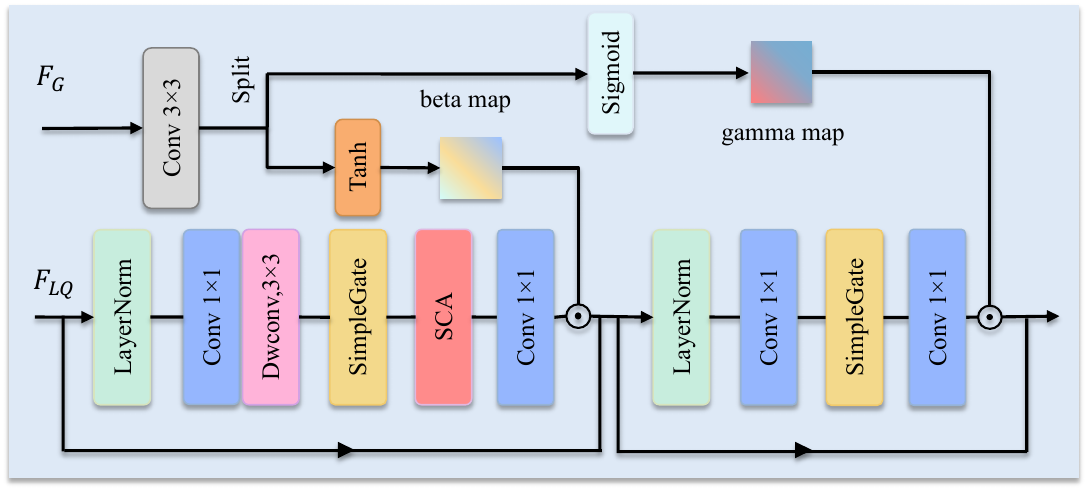}
  \caption{The proposed Guided Color Enhancement Block.}
  \label{fig:figure5}
\end{figure*}

\subsubsection{Texture Transfer.}
\label{subsubsec:texture_transfer.}
Given the enhanced feature obtained from the previous color-guidance stage and the reference feature extracted from the high-quality cover image, our next goal is to reconstruct high-frequency spatial details for the target frame. To align local structures, $\bm{F}_{EN}$ is divided into non-overlapping patches of size $p\times p$, and each patch $\bm{P}_{EN}^{(i)}$ searches for its $k$ most similar patches in $\bm{F}_{Ref}$ (where $k=3$). The top-$k$ match indices $\mathcal{J}^*$ are determined by normalized cross-correlation:$$\mathcal{J}^* = \mathop{\text{TopK}_j}
\frac{\langle \bm{P}_{EN}^{(i)}, \bm{P}_{Ref}^{(j)} \rangle}
{\|\bm{P}_{EN}^{(i)}\|_2 \, \|\bm{P}_{Ref}^{(j)}\|_2}.$$ To reduce computational cost, both features are downsampled by a ratio $r$ before matching, and the resulting indices are remapped to the original coordinate grid. Since both the feature size and patch size shrink by $r$, the complexity decreases approximately by $1/r^4$. The corresponding top-3 matched reference patches are then gathered to form the local reference feature space. After structural alignment, we refine $\bm{F}_{EN}$ using a \textit{Similar Window Cross Attention} (SWCA) operation. For each local query window in $\bm{F}_{EN}$, we compute multi-head cross-attention where the query tokens $\bm{Q}$ are derived from $\bm{F}_{EN}$, and the key--value tokens $\bm{K}, \bm{V}$ are extracted from the concatenated tokens of the corresponding $k=3$ most similar reference patches. Prior to the attention calculation, $\bm{Q}$ and $\bm{K}$ are $\ell_2$-normalized along the channel dimension so that the attention is computed based on cosine similarity, which improves robustness to local contrast and illumination variations. The aggregated result of cross-attention is denoted as $\bm{Y}$. To prevent over-reliance on the reference, we adopt a conservative residual fusion strategy: the input feature $\bm{F}_{EN}$ is linearly projected to $\bm{X}_{base}$, and the super-resolved feature $\bm{F}_{SR}$ is obtained by a gated interpolation between $\bm{X}_{base}$ and $\bm{Y}$:$$\bm{F}_{SR} = (1-\alpha)\bm{X}_{base} + \alpha\bm{Y},$$where $\alpha\in[0,1]$ is a learnable coefficient that adaptively balances the contribution of the reference. This mechanism preserves the structural integrity of $\bm{F}_{EN}$ while injecting high-frequency texture details from $\bm{F}_{Ref}$.

\subsection{Feature Fusion and Reconstruction.}
Finally, the feature $\bm{F}_{SR}$ is fused with the input feature through convolution and refined by several Swin Transformer Layers (denoted as $\mathrm{STL}(\cdot)$):
\begin{equation}
\bm{F}_{out} = \mathrm{STL}\!\left(
\text{Conv}\!\left([\bm{F}_{EN},\,\bm{F}_{SR}]\right)
\right),
\end{equation}
An upsampling head then reconstructs the high-resolution output, producing the final $4\times$ super-resolved image:
\begin{equation}
\bm{X}_{SR} = \text{UpSample}_{\times 4}\!\left(\bm{F}_{out}\right),
\end{equation}

\subsection{Loss Function}
During training, we aim for the input image to recover both the color and texture of the ground truth. To achieve this, we employ two losses — a color loss and a reconstruction loss — applied at different stages of the network. The overall loss is :
\begin{equation}
\mathcal{L}=\mathcal{L}_{color}+\mathcal{L}_{rec},
\end{equation}

\noindent\textbf{Color loss.}
To minimize the influence of color inconsistencies before performing super-resolution, we adopt a loss function between $\bm{X}_{EN}$ and $\bm{X}_{GT\downarrow}$, which is similar to \cite{lou2025learning},
\begin{equation}
\mathcal{L}_{color}=\lambda_1\mathcal{L}_{2}+\lambda_2\mathcal{L}_{ssim}+\lambda_3\mathcal{L}_{vgg},
\end{equation}
where $\lambda_1$, $\lambda_2$, $\lambda_3$ are set to 10, 0.5, and 0.005.
 
\noindent\textbf{Reconstruction loss.}
We adopt the commonly used $\mathcal{L}_1$ loss between $\bm{X}_{SR}$ and $\bm{X}_{GT}$ as reconstruction loss.

 \section{Experiments}
\label{sec:Experiments}

\subsection{Evaluation}

\begin{table}[t]
  \centering
  \small
  \caption{Quantitative comparison with representative methods on two settings of Live2K dataset. The best results are marked in \textbf{bold}. “/” are absent results due to the unavailable code.}
  \label{tab:table4}
  \resizebox{\textwidth}{!}{
  \begin{tabular}{l|l|c|cccc|cccc}
    \hline
    \multirow{2}{*}{\textbf{Category}} 
    & \multirow{2}{*}{\textbf{Methods}} 
    & \multirow{2}{*}{\textbf{\#Params}} 
    & \multicolumn{4}{c|}{\textbf{iPhone}} 
    & \multicolumn{4}{c}{\textbf{OPPO}} \\
    \cline{4-11}
    & & 
    & PSNR $\uparrow$ 
    & SSIM $\uparrow$ 
    & LPIPS $\downarrow$ 
    & $\Delta E \downarrow$ 
    & PSNR $\uparrow$ 
    & SSIM $\uparrow$ 
    & LPIPS $\downarrow$ 
    & $\Delta E \downarrow$ \\
    \hline

    SISR 
    & SwinIR \cite{liang2021swinir} 
    & 8.05M 
    & 26.58 & 0.8447 & 0.2353 & 4.21 & 27.38 & 0.8543 & 0.2176 & 4.14 \\
    \hline

    \multirow{4}{*}{BurstSR} 
    & BIPNet \cite{dudhane2022burst} 
    & 1.76M 
    & 25.97 & 0.8204 & 0.2618 & 4.78 & 26.93 & 0.8398 & 0.2630 & 4.67 \\

    & Burstormer \cite{dudhane2023burstormer} 
    & 1.56M 
    & 26.03 & 0.8205 & 0.2605 & 4.77 & 26.66 & 0.8311 & 0.2668 & 4.83 \\

    & BurstM \cite{kang2024burstm} 
    & 13.4M 
    & 26.12 & 0.8262 & 0.2545 & 4.76 & 26.31 & 0.8246 & 0.2514 & 4.99 \\

    & QMambaBSR \cite{di2025qmambabsr} 
    & / 
    & / & / & / & / 
    & / & / & / & / \\
    \hline

    \multirow{3}{*}{RefSR} 
    & ERVSR \cite{kim2023efficient} 
    & 5.76M & 25.40 & 0.8190 & 0.2695 & 4.94 
    & 26.59 & 0.8195 & 0.2708 & 5.06 \\
    
    & DATSR \cite{cao2022reference} 
    & 5.58M 
    & 26.34 & 0.8407 & 0.2492 & 4.50 & 26.64 & 0.8445 & 0.2422 & 4.68 \\

    & MRefSR \cite{zhang2023lmr} 
    & 11.8M 
    & 26.34 & 0.8372 & 0.2419 & 4.44 & 26.97 & 0.8475 & 0.2243 & 4.47 \\

    & RebaIR \cite{bernasconi2025rebair} 
    & 72.1M 
    & / & / & / & / 
    & / & / & / & / \\
    \hline

    & \textbf{Ours} 
    & 5.06M 
    & \textbf{26.97} & \textbf{0.8497} & \textbf{0.2268} & \textbf{4.05} 
    & \textbf{27.78} & \textbf{0.8657} & \textbf{0.2005} & \textbf{3.95} \\
    \hline
  \end{tabular}
  }
\end{table}

During the testing phase, we perform inference on full-resolution images to address the re-selection and enhancement task of Live Photo under real-world conditions.
Moreover, due to the presence of the guidance image, patch-based inference is not applicable in our setting. We evaluate our method using PSNR and SSIM metrics, both of which are computed on the Y channel (luminance) of the YCbCr color space.

\subsection{Implementation Details}
Since the super-resolution scale factors of both settings are non-integer, we first upsample the input image to half the spatial size of the guidance image using interpolation within the network. We then apply PixelUnshuffle to perform a 2× downsampling, followed by a 4× super-resolution reconstruction in the final stage. The detailed parameters of our network can be found in the supplementary material.

We employ the Muon optimizer~\cite{jordan2024muon} with auxiliary Adam~\cite{kingma2014adam} for bias and gain parameters. The matrix weights are updated with Muon using a 10$\times$ higher learning rate, while the scalar parameters use Adam ($\beta_1{=}0.9$, $\beta_2{=}0.99$).  We augment the
 training data with random horizontal and vertical flipping or different random
 rotations of $90^\circ,\, 180^\circ,\, \text{and}\, 270^\circ$. The model is trained with a batch size of 24 and an initial learning rate of 2e-4, which is cosine-annealed to 4e-6 over 200K iterations. All experiments are conducted on two NVIDIA RTX 5090 GPUs.

\subsection{Comparisons with Existing Methods}

For comparison, we selected state-of-the-art methods from the domains of Single Image Super-Resolution (SISR), Reference-based Super-Resolution (RefSR), and Burst Image Super-Resolution (BSR). All models are retrained on our Live2K dataset for a fair comparison. We uniformly upsample the input images to half the size of the GT, then apply PixelUnshuffle for a further 2× downsampling before feeding them into the network. Crucially, performing direct inference on 4K images is computationally infeasible for almost all existing methods, as the massive memory footprint strictly exceeds standard GPU limits. To make the evaluation even possible, we were compelled to scale down these architectures. For example, we discarded the multi-scale structures in DATSR and MRefSR, retaining only operations at the lowest scale, and reduced the base channel number of BIPNet from 64 to 32. Details regarding modifications to other models are provided in the supplementary material. Other training settings remain consistent with the original implementations.

\begin{figure*}[t]
  \centering
  \includegraphics[width=\textwidth]{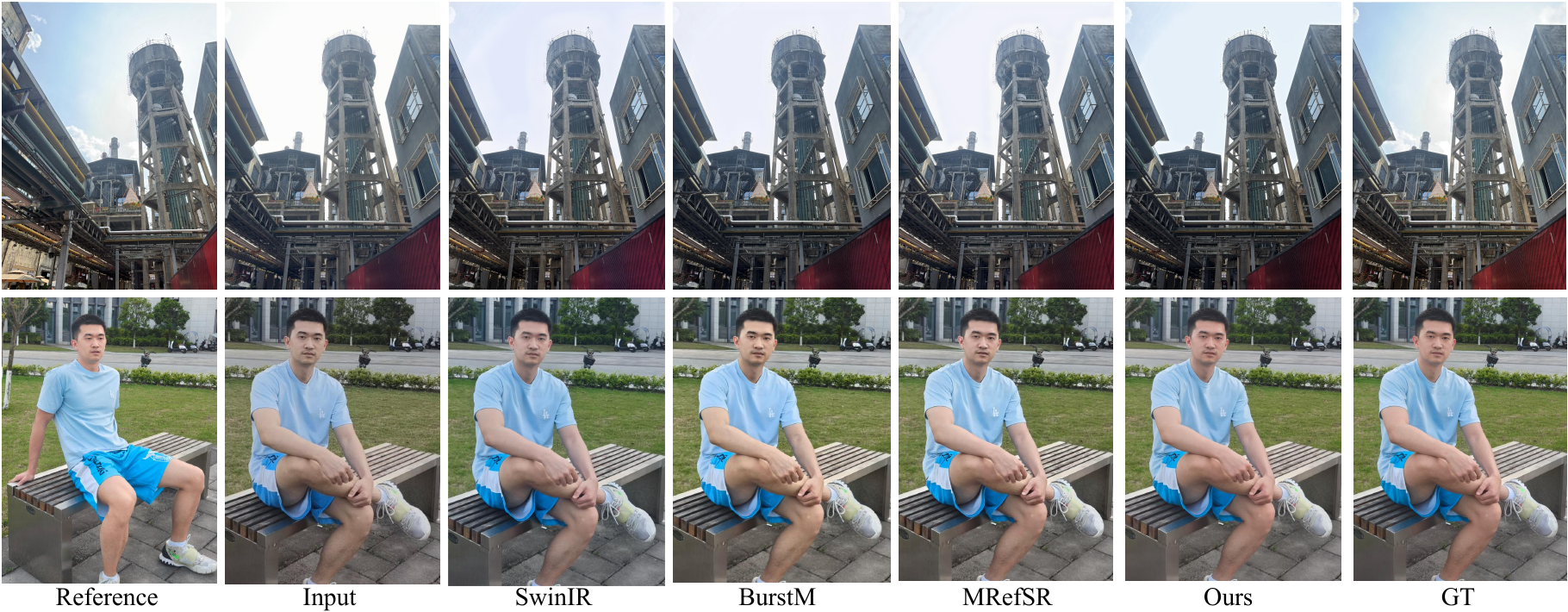}
  \caption{Visual comparison of overall color fidelity with SOTA methods }
  \label{fig:figure6}
\end{figure*}

\begin{figure*}[t]
  \centering
  \includegraphics[width=\textwidth]{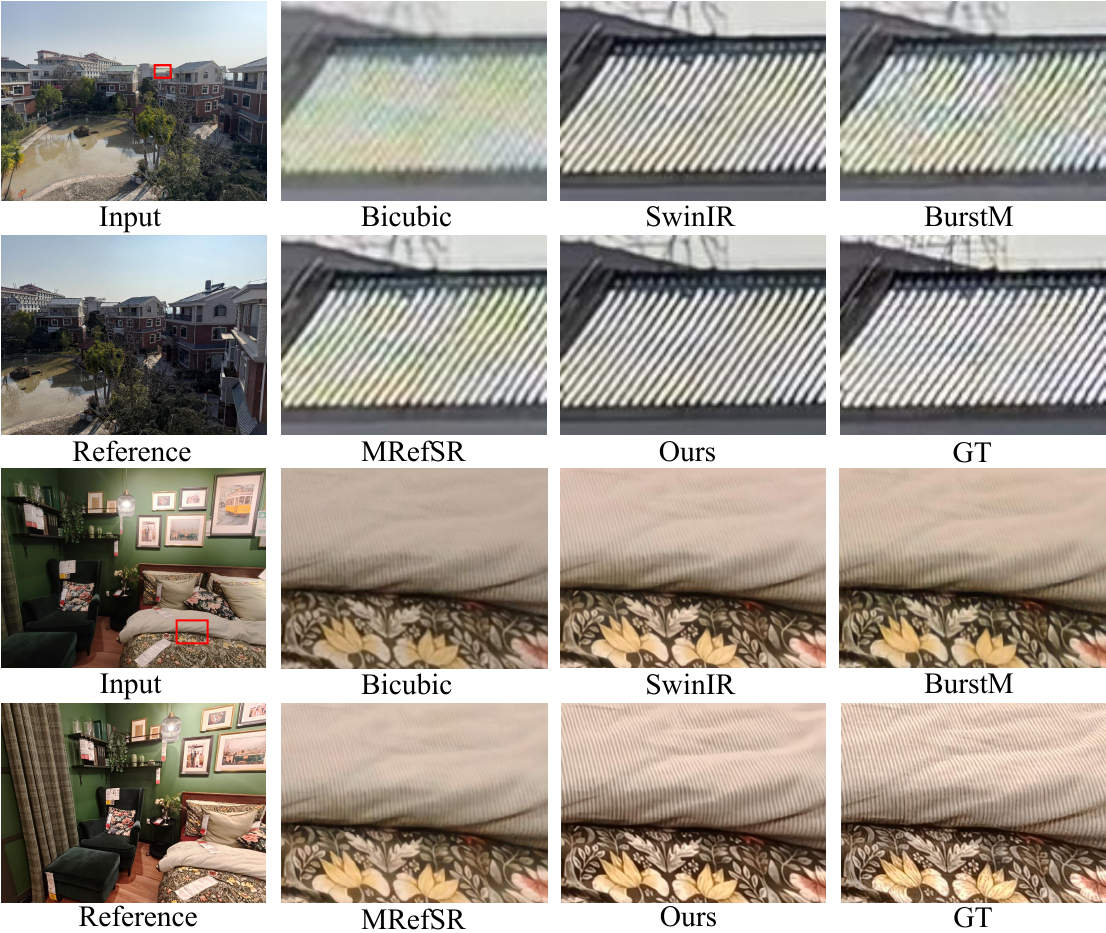}
  \caption{Visual comparison of texture with SOTA methods. }
  \label{fig:figure7}
\end{figure*}

As shown in Tab.~\ref{tab:table4}, our method achieves a significant improvement over previous approaches on both subcategories of the dataset. We also provide visual examples to demonstrate the comparisons in terms of overall color and local details, as presented in Fig.~\ref{fig:figure6} and Fig.~\ref{fig:figure7}. These images clearly demonstrate our advantage in recovering textures and color from severely degraded video frames. Notably, the input image in the first row of Fig~\ref{fig:figure6} is severely overexposed, resulting in an almost purely white sky. Although our method recovers some colors compared to other models, the result is still suboptimal, indicating that there is room for improvement in our utilization of the guidance image.

\subsection{Ablation Study}

\noindent\textbf{Effects of each component.} To verify the effectiveness of each component in our network, we conduct an ablation study on the OPPO subset by progressively removing modules from the full model. The last configuration represents our complete model. 
In the third experiment, we removed the reference image, completely discarded the detail transfer module, and employed the basic NAFBlocks in the color modulation module. In the second experiment, we ablated the intermediate color supervision loss. In the first experiment, we discarded the multi-frame inputs, utilizing only a few convolutional layers to extract features from the input image.
As illustrated in Tab.~\ref{tab:ablation1}, introducing multi-frame information yields an improvement of 0.29 dB. Adding the color supervision loss brings a 0.06 dB gain, although marginal, it introduces almost no additional inference burden. Furthermore, incorporating the reference image results in a 0.3 dB boost.

\begin{table*}[htbp] % 如果你是双栏排版并希望并排后的表格跨越两栏，请保留 table*；如果是单栏，改为 table 即可
\centering

% ----------------- 第一个表格 -----------------
\begin{minipage}{0.48\linewidth} % 占据页面 48% 的宽度
  \centering
  \setlength{\tabcolsep}{4pt}
  \renewcommand{\arraystretch}{1.0}
  \caption{Ablation study on each component of our framework.}
  \label{tab:ablation1}
  \resizebox{\linewidth}{!}{% 将表格缩放至 minipage 的宽度
  \begin{tabular}{c c c | c c}
  \hline
  \multirow{2}{*}{Multi-Frame} & \multirow{2}{*}{Color Loss} & \multirow{2}{*}{Reference} & \multicolumn{2}{c}{OPPO} \\
   &  &  & PSNR & SSIM \\
  \hline
             &            &            & 27.13 & 0.8516 \\
  \checkmark &            &            & 27.42 & 0.8596 \\
  \checkmark & \checkmark &            & 27.48 & 0.8627 \\
  \checkmark & \checkmark & \checkmark & 27.78 & 0.8657 \\
  \hline
  \end{tabular}%
  }
\end{minipage}\hfill % \hfill 用于在两个 minipage 之间自动填充空白，将它们推向两端
% ----------------- 第二个表格 -----------------
\begin{minipage}{0.48\linewidth} % 占据页面 48% 的宽度
  \centering
  \setlength{\tabcolsep}{4pt}
  \renewcommand{\arraystretch}{1.0}
  \caption{Ablation study on two components within the Guided Enhancement module.}
  \label{tab:ablation2}
  \resizebox{\linewidth}{!}{% 将表格缩放至 minipage 的宽度
  \begin{tabular}{c c | c c}
  \hline
  \multirow{2}{*}{Color Modulation} & \multirow{2}{*}{Texture Transfer} & 
  \multicolumn{2}{c}{OPPO} \\
   &  &  PSNR & SSIM \\
  \hline
             &            & 27.48 & 0.8627 \\
  \checkmark &            & 27.66 & 0.8643 \\
             & \checkmark & 27.68 & 0.8644 \\
  \checkmark & \checkmark & 27.78 & 0.8657 \\
  \hline
  \end{tabular}%
  }
\end{minipage}

\end{table*}

\noindent\textbf{Effects of designs about reference.} We conduct experiments to verify the effectiveness of two components in the Guided Enhancement module. As shown in Tab.~\ref{tab:ablation2}, using either color modulation or texture transfer independently yields a PSNR improvement of approximately 0.2 dB, while combining them results in a total gain of 0.3 dB. This demonstrates that the two components are not only individually effective but also highly complementary, working synergistically to achieve better reconstruction performance.

\subsection{Computational cost Comparison}
We evaluate the GPU memory consumption and inference speed of models at a ground truth image resolution of 4096×3072, as listed in Tab.~\ref{tab:runtime_memory}. The inference speed is reported as the average over 100 test images. The results show that our model achieves the fastest inference speed with a relatively moderate memory footprint.
\begin{table}[htbp]
\centering
\setlength{\tabcolsep}{4pt}
\renewcommand{\arraystretch}{1.0}
\caption{Comparison of GPU memory and runtime on 4K images.}
\resizebox{0.8\columnwidth}{!}{%
\begin{tabular}{l | c c c c c c}
\hline
Model & SwinIR & BIPNet & BurstM & DATSR & MRefSR & LPENet \\
\hline
Memory (GB) & 16.25 & 26.93 & 17.47 & 24.65 & 27.58 & 21.36 \\
Runtime (s)     & 1.419 & 1.036 & 1.80 & 21.05 & 34.86 & 0.865 \\
\hline
\end{tabular}
}
\label{tab:runtime_memory}
\end{table}

\subsection{Real-World Evaluation}
\begin{figure*}[t]
  \centering
  \includegraphics[width=\textwidth]{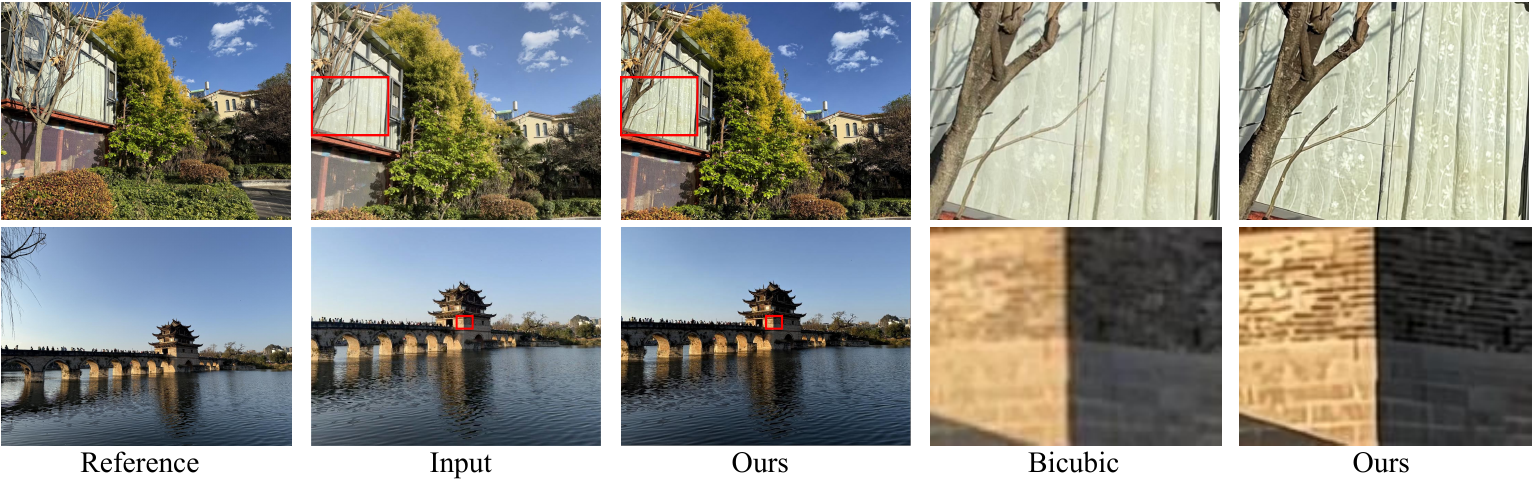}
  \caption{Visual results of our model in real-world condition.}
  \label{fig:figure8}
\end{figure*}

The reference images and input frames in our dataset originate from distinct Live Photos, in real-world applications, the cover photo of a Live Photo naturally serves as the reference image. To evaluate our model's performance in this practical setting, we captured Live Photos with significant camera movements, utilizing the cover image as guidance and the final 9 video frames as inputs. Ground Truth is unavailable in such real-world scenarios, we solely provide qualitative visual comparisons. As shown in Fig.~\ref{fig:figure8}, our approach consistently enhances the visual quality of the re-selected frames.

\subsection{Cross-device Generalization}
% Due to significant variations in ISP pipelines across manufacturers, applying a universal model is impractical, making device-specific training the industry norm. Reflecting this, our sub-datasets exhibit distinct degradation profiles (e.g., noise, color, and scale factors), necessitating separate training. To explore cross-device generalization, we fine-tuned the OPPO-pretrained model for 5,000 iterations using only 60 images from the iPhone training set. To explore the potential of cross-device generalization, we fine-tuned the model pre-trained on the OPPO dataset for 5,000 iterations, using only the first 60 images from the iPhone training dataset. When evaluated on the iPhone test set, as shown in Tab~\ref{tab:psnr_ssim_comparison} and Fig~\ref{fig:figure9}, this minimal fine-tuning yields substantial quantitative and visual improvements, demonstrating our model's potential for cross-device transferability.

\begin{figure*}[htbp] 
    \centering
    % 左侧：表格
    \begin{minipage}[t]{0.23\textwidth}
        % \vspace{-0.3cm} 
        \centering
        \captionof{table}{Quantitative comparison on iPhone dataset.}
        \label{tab:psnr_ssim_comparison}
        \resizebox{\linewidth}{!}{%
        \begin{tabular}{lcc}
            \toprule
            \textbf{Setting} & \textbf{PSNR} & \textbf{SSIM} \\
            \midrule
            w/o fine-tuned & 20.63 & 0.6769 \\
            w/ fine-tuned  & 25.43 & 0.8132 \\
            \bottomrule
        \end{tabular}%
        }
    \end{minipage}
    \hfill
    % 右侧：图片
    \begin{minipage}[t]{0.72\textwidth}
        \vspace{0pt} 
        \centering
        \includegraphics[width=\linewidth]{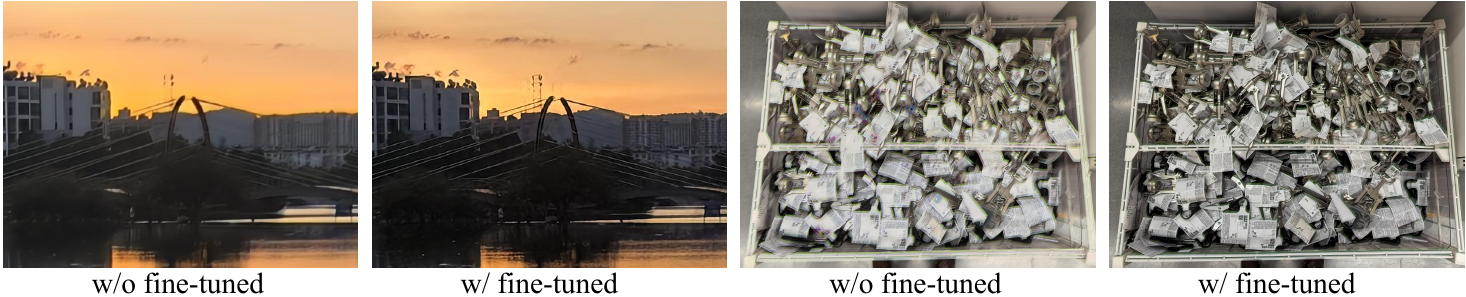}
        \captionof{figure}{Visual performance comparison on iPhone dataset.}
        \label{fig:figure9}
    \end{minipage}
\end{figure*}

The ISP hardware pipelines and algorithms vary significantly across different manufacturers, making it impractical to directly apply a universal model to diverse devices. Consequently, device-specific training is commonly adopted in the industry. True zero-shot ISP invariance remains challenging in industrial mobile imaging systems, and our method does not fully solve this problem. Reflecting this, our sub-datasets exhibit distinct degradation profiles, such as noise, color, and scale factors, necessitating separate training. To explore the potential of cross-device generalization, we fine-tuned the model pre-trained on the OPPO dataset for 5,000 iterations using only the first 60 images from the iPhone training set. When evaluated on the iPhone test set, as shown in Tab.~\ref{tab:psnr_ssim_comparison} and Fig.~\ref{fig:figure9}, this minimal fine-tuning yields substantial quantitative and visual improvements, demonstrating the potential of our model for cross-device transferability.

\section{Limitation}
Our dataset is mainly limited to well-lit daytime scenes, lacking night-time data. Capturing Live Photos in low-light environments naturally introduces longer exposure times, severe noise, dropped frames, and significant inter-frame motion blur. Because these degradations differ fundamentally from daytime data, our trained model may not generalize well to night scenes. Future research may focus on building a universal dataset or a domain-specific dataset tailored for extreme low-light scenarios, to effectively bridge this domain gap and ensure robust performance across all lighting conditions.

\section{Conclusion}
\label{sec:conclusion}

In this work, we introduced the Live Photo Cover Frame Reselection and Enhancement (LPRE) task, addressing the quality gap between the still cover and the accompanying video frames in Live Photos, with the goal of improving the visual quality of the re-selected cover frame. We analyzed the fundamental causes of this discrepancy from the perspective of the imaging pipeline and demonstrated that the issue cannot be resolved within the system itself. To facilitate research in this direction, we constructed Live2K, the first large-scale real-world dataset specifically tailored for the LPRE task. Furthermore, we proposed a unified one-stage baseline model that jointly performs multi-frame fusion, guided enhancement, and super-resolution within a single framework. Extensive experiments on Live2K, alongside evaluations in real-world scenarios, validate the effectiveness and robustness of our approach, establishing a solid benchmark for future studies.

\noindent\textbf{Acknowledgment.} This work was supported by National Natural Science Foundation of China (No.62476051) and Sichuan Natural Science Foundation (No.2024NSFTD0041).

% TODO FINAL: This \clearpage needs to be removed from both review and camera-ready versions.

\bibliographystyle{splncs04}
\bibliography{main}
\end{document}